\documentclass[10pt,twocolumn,letterpaper]{article}

\usepackage{iccv}
\usepackage{times}
\usepackage{epsfig}
\usepackage{graphicx}
\usepackage{amsmath}
\usepackage{amssymb}

\usepackage{subfig}

\usepackage[pagebackref=true,breaklinks=true,letterpaper=true,colorlinks,bookmarks=false]{hyperref}

 \iccvfinalcopy 

\def\iccvPaperID{1191} 

\ificcvfinal\fi
\begin{document}

\title{EML-NET:An Expandable Multi-Layer NETwork for Saliency Prediction}

\author{
Sen Jia\\
Ryerson University\\
{\tt\small sen.jia@ryerson.ca}
\and
Neil D. B. Bruce\\
Ryerson University\\
Vector Institute\\
{\tt\small bruce@ryerson.ca}
}

\maketitle

\begin{abstract}
Saliency prediction can benefit from training that involves scene understanding that may be tangential to the central task; this may include understanding places, spatial layout, objects or involve different datasets and their bias. One can combine models, but to do this in a sophisticated manner can be complex, and also result in unwieldy networks or produce competing objectives that are hard to balance. In this paper, we propose a scalable system to leverage multiple powerful deep CNN models to better extract visual features for saliency prediction. Our design differs from previous studies in that the whole system is trained in an almost end-to-end piece-wise fashion. The encoder and decoder components are separately trained to deal with complexity tied to the computational paradigm and required space. Furthermore, the encoder can contain more than one CNN model to extract features, and models can have different architectures or be pre-trained on different datasets. This parallel design yields a better computational paradigm overcoming limits to the variety of information or inference that can be combined at the encoder stage towards deeper networks and a more powerful encoding. Our network can be easily expanded almost without any additional cost, and other pre-trained CNN models can be incorporated availing a wider range of visual knowledge. We denote our expandable multi-layer network as EML-NET and our method achieves the state-of-the-art results on the public saliency benchmarks, SALICON, MIT300 and CAT2000.
\end{abstract}

\section{Introduction}

\begin{figure}[t]
\centering
\includegraphics[width=0.5\textwidth]{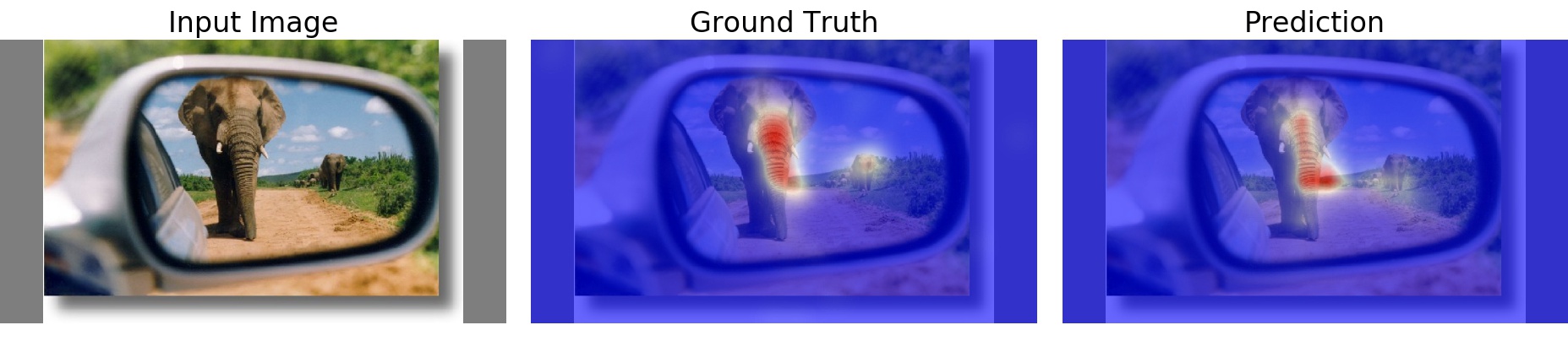}\\
\includegraphics[width=0.5\textwidth]{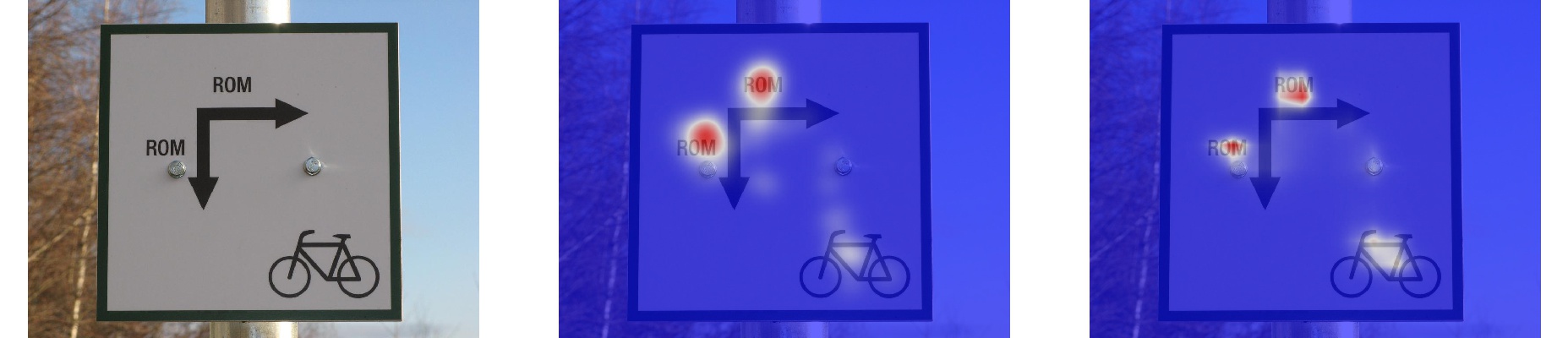}\\
\vspace{.5em}
\includegraphics[width=0.5\textwidth]{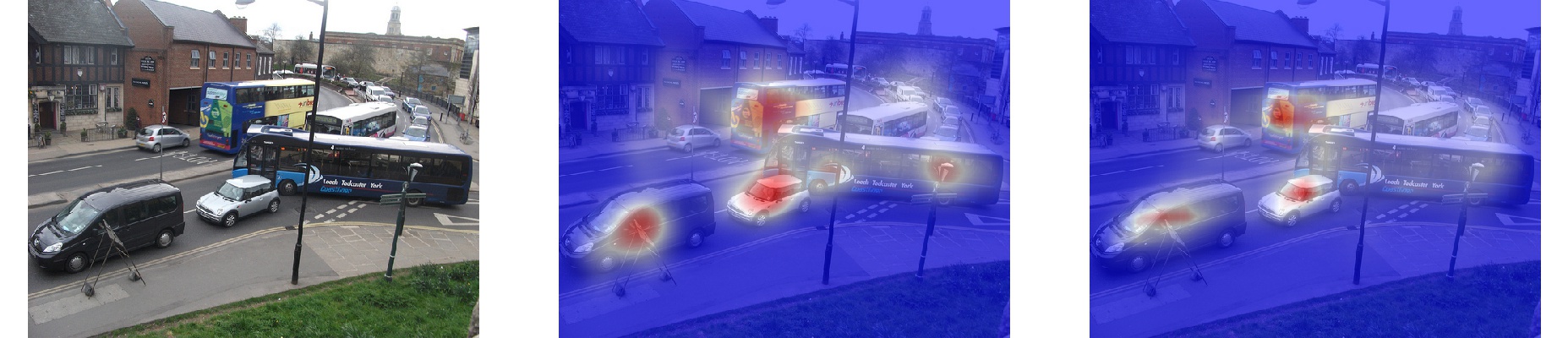}
\caption{Examples of predicted saliency maps. First row : CAT2000; Second row : MIT1003; Third row : SALICON.}
\label{fig:examples}
\end{figure}

In looking at an image, humans tend to fixate on Regions Of Interest (ROIs) and less salient parts of the image are ignored. This attention mechanism of the Human Visual System (HVS) plays an important role in vision tasks such as object classification \cite{Gao04}, video analysis \cite{Yao17,Zhang17a}, image compression\cite{Yu09}, action classification \cite{Sharma12} and quality assessment \cite{Jia17a,Jia17b}. Simulating where humans gaze in pictures in computer vision is referred to as visual saliency prediction, as shown in Figure~\ref{fig:examples}. The stimulus-selectivity of the HVS can be categorized into two modes: top-down and bottom-up \cite{Connor04}. The former requires that observers have prior knowledge of a task or contextual information about the picture to predict ROIs. The latter considers the is data-driven and the content of an image ignoring these external factors determines where humans may gaze at. 

To predict salient regions using computers, extensive studies have been carried out with the intent to capture discriminative visual features \cite{Koch87,Itti98,Bruce05,Valenti09,Erdem13}. With the development of Convolutional Neural Networks (CNNs), better  representations of visual features can be extracted in an end-to-end design. A common paradigm involves applying a CNN model that is pre-trained on the ImageNet dataset \cite{ImageNet} in saliency applications \cite{DeepGaze1,DeepGaze2,Wang17a,MLNet,DSCLRCN,DeepFix, PDP,SAMres,Salicon,EDN,iSEEL,SalGAN}. The success behind these approaches suggests that learned visual features from other domains can be transferred successfully to the task of saliency prediction.

Many CNN architectures have been proposed in the area of image classification \cite{AlexNet,VGGNet,ResNet,DenseNet,NasNet}, where a deep CNN model is able to achieve a higher accuracy for classification \footnote{In this paper, a deeper or more powerful CNN model refers to one that produces higher classification accuracy on the ImageNet dataset.}. However, CNN models used in state-of-the-art saliency applications are relatively shallow, such as VGGNet-16 \cite{DeepGaze2,Wang17a,MLNet,PDP,DeepFix,iSEEL,SalGAN} or ResNet-50 \cite{DSCLRCN,SAMres}. In the work of \cite{Salicon}, the deeper model, GoogleNet, did not achieve better performance due to the limited training set. With the availability of large-scale saliency data, this issue is no longer a central problem in applying deep CNN models. The SALICON \cite{SALICONdb} dataset is used in this work, which contains $10,000$ annotated images for training.

Required computational space is another bottleneck to applying deep CNN models. Compared with image classification, CNN models used for saliency prediction require less parameters because the fully connected layer is replaced by convolutional layers (considered as a decoder). However, more space is required to represent the data due to larger input sizes. This problem becomes even more problematic when extracting features from multiple layers \cite{DeepGaze2,MLNet,Wang17a} or when involving a recurrent network \cite{DSCLRCN,SAMres}. The number of parameters of a decoder grows linearly with the number of connected intermediate convolutional layers. 

Recall that the CNN model applied to saliency prediction contains prior knowledge about objects. But the information concerning objects is still limited because there are only $1,000$ categories (some very specific) in the ImageNet dataset. It is still a challenge of how to introduce visual information from other domains for saliency prediction, such as face detection \cite{Cerf07} or a combination of multiple vision domains \cite{Borji12}. Doing so further increases complexity, may involve balancing objectives, and implies a greater requirement on space when combining multiple CNN models which have been pre-trained for different visual tasks.

To utilize a more powerful encoder, in this work, we propose a scalable system for saliency prediction. Our method differs from previous studies in that the whole system is split into small modules and each module is trained separately instead of joint training. Not only are the encoder and the decoder trained separately, but also each CNN model in the encoder stage is trained separately. This modular design significantly reduces the required computational complexity and required space allowing effective use of deep state-of-the-art CNN models such as NasNet \cite{NasNet} and DenseNet \cite{DenseNet}, to extract deeper representations of saliency. Moreover, due to the scalability of this method, multiple deep CNN models can be combined in the encoder stage to deliver `wider' prior visual knowledge from different tasks, such as object \cite{ImageNet} and scene \cite{PLACE} recognition. The decoder component needs training from scratch whenever a new model is added to the encoder stage, but the computational cost to doing so is very small (see Section~\ref{sec:decoder}). In this paper, we define our proposed saliency system as Expandable Multi-Layer NETwork, EML-NET.

Another contribution of our paper is that we modify existing saliency metrics to formulate a combined loss function. The saliency metrics of Pearson's Correlation Coefficient (CC) and Normalized Scanpath Saliency (NSS) are modified into the representation of dissimilarity, and they are combined with the metric of Kullback-Leibler Divergence (KLD) to compute loss for our model, see Section~\ref{sec:loss}. Given disagreement among the most suitable metric, this allows for multiple competing objectives to be satisfied in concert. We train our EML-NET on the large saliency dataset, SALICON, and evaluate on three public saliency benchmarks, SALICON \cite{SALICONdb}, MIT300 \cite{MIT300a, MIT300b} and the CAT2000 dataset \cite{CAT2000}. Our experimental results show that the proposed EML-NET outperforms other state-of-the-art saliency algorithms, including the ResNet-50 combined with a recurrent network \cite{SAMres,DSCLRCN}; this is especially so in considering the NSS metric, see Section~\ref{sec:results}. Our design can potentially be combined with other saliency algorithms, e.g. multi-resolution \cite{Salicon} or recurrent models, see Section~\ref{sec:related}.

\section{Related Work}\label{sec:related}
Koch and Ullman studied saliency maps using a neural network and a set of basic visual features \cite{Koch87}. Their work later inspired the first computational bottom-up saliency method \cite{Itti98}, in which Itti et al. extracted 42 feature maps including colour, intensity and orientation features. Bruce et al. \cite{Bruce05} applied independent component analysis on image patches to compute saliency maps. Valenti \cite{Valenti09} et al. proposed to utilize the information of colour and edges for saliency prediction. Erdem et al. \cite{Erdem13} investigated the contribution of each visual feature using region co-variance matrices. Zhang et al \cite{BMS} applied thresholds in the CIE Lab colour space to compute boolean maps and subsequently computing saliency maps. 

The use of CNNs has proven to be very powerful to extract discriminative features in the area of image classification \cite{AlexNet,VGGNet,ResNet,NasNet,Jia16a,Jia16b}. Applying CNNs for saliency prediction was first proposed by Vig et al. \cite{EDN}, in which they combined information from multiple layers to train a linear classifier. The work of \cite{DeepGaze1} has shown that the visual features learned from object recognition \cite{AlexNet} can be used to predict saliency maps. Therefore, to better utilize off-the-shelf image features, deeper network architectures have been applied for saliency prediction. For instance, the VGGNet-16 model \cite{VGGNet} was used in application to saliency prediction \cite{DeepFix,DeepGaze2,PDP,MLNet,Wang17a,iSEEL,SalGAN}. 

More recently, the combination of ResNet-50 \cite{ResNet} with a recurrent network has been proposed \cite{DSCLRCN,SAMres} and their methods achieved state-of-the-art results on the MIT300 \cite{MIT300a} benchmark. Similar to the work of \cite{SAMres}, we also use multiple saliency metrics to formulate a combined loss function, however, we modify the metrics of NSS and CC into the representation of dissimilarity without using empirical coefficients, see Section~\ref{sec:loss}. The work of \cite{DSCLRCN} also exploited features from scene recognition, the PLACE-CNN \cite{Zhou14} model. However, their design is still based on jointly training the whole system. Therefore, it is difficult to extend their system to include features or modes of inference captured by other models. More importantly, the input image has to be resized to adjust to the place model because a linear layer of $128$ units is applied to output scene features. This process results in a tremendous number of parameters, see Section~\ref{sec:encoder}, and may cause a loss of information in both the input and output spaces. 

All the CNN models used in saliency applications are relatively `shallow' (lower classification accuracy on ImageNet) compared with the state-of-the-art models in object recognition, e.g. DenseNet \cite{DenseNet} or NasNet \cite{NasNet}. Huang et al. \cite{Salicon} fine-tuned different CNN architectures, AlexNet, VGGNet-16 and GoogleNet \cite{GoogleNet} (all pre-trained on ImageNet) at different image scales for saliency prediction. The GoogleNet model outperforms VGGNet-16 in the task of image classification. However, there was no gain provided by GoogleNet in their comparisons. The reason behind this could be that the training set they used is too small ($450$ images from the OSIE dataset \cite{OSIE}). This problem is partially solved by the recently published large-scale saliency dataset, SALICON \cite{SALICONdb}. The size of images from SALICON is $640 \times 480$, almost five times larger than the images in ImageNet. Moreover, the decoder needs extra parameters to utilize multi-layer features \cite{MLNet,Wang17a,DeepGaze1,DeepGaze2}.

Our EML-NET is trained in an almost end-to-end design so that deep CNN models can be applied on large images to extract multi-layer features. The powerful encoder consists of NasNet from ImageNet and DenseNet from PLACE365 \cite{PLACE}. Our method can be considered as an alternative to the design \cite{DSCLRCN,SAMres} of a simple encoder(ResNet-50) combined with a complex decoder (long short-term memory). Moreover, EML-NET could potentially be combined with a recurrent network or models from other tasks \cite{Cerf07,Borji12}.

\begin{figure*}[tph]
\centering
\subfloat{\includegraphics[width=1\textwidth]{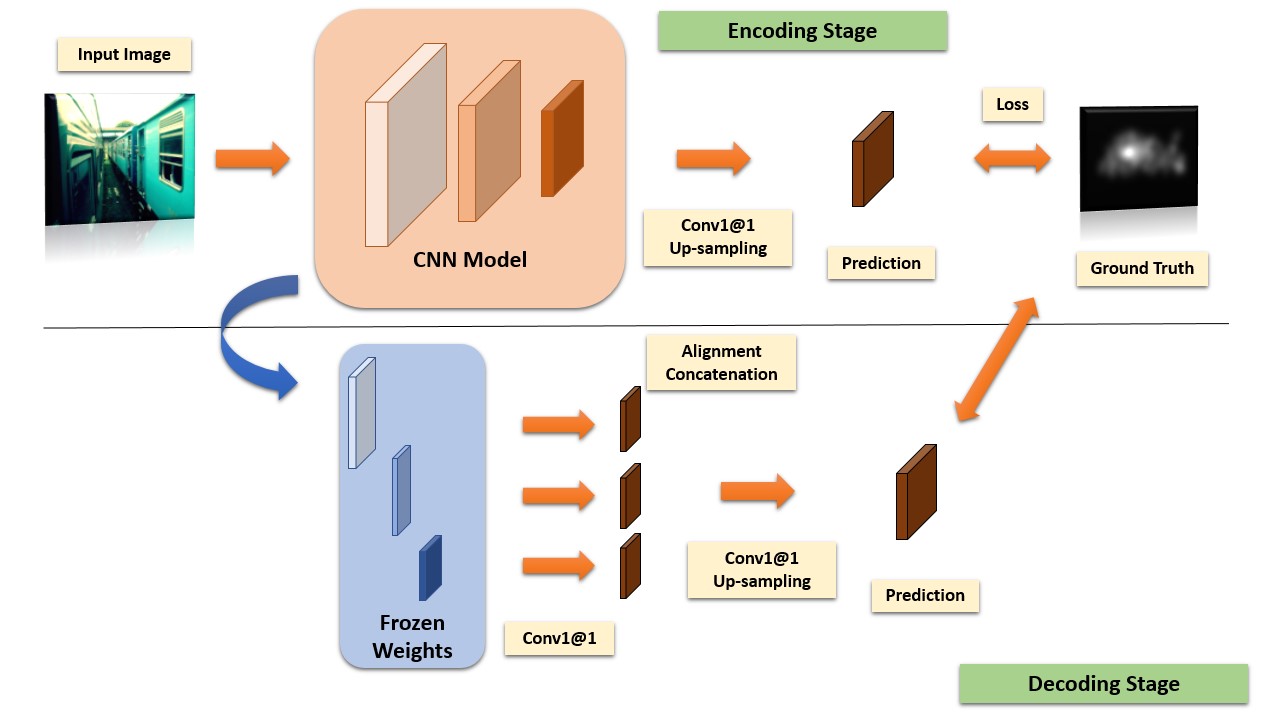}}
\caption{Flowchart of our EML network.}
\label{fig:eml}
\end{figure*}

\section{Expandable Saliency System}
\label{sec:EML-NET}

Consider $\vec{x_i}$ as the $i$th image in the training set and $\vec{y_i}$ is the associated ground truth, $\{\vec{x_i}, \vec{y_i}\} \in \Re^{n}$. The convolutional layers of a CNN model map an input image into the feature space via a sequence of operations, $E(\vec{x_i}) \rightarrow \vec{a_i} \in \Re^{m},\ E(\cdot)=W^L(W^{L-1}(\cdots (W^1(x_i))))$, where $W^L$ is the weight of the $L$th layer. This mapping process $E(\cdot)$ is considered to be an encoder stage to extract features $\vec{a_i}$. Then a decoder module is applied to up-sample and predict saliency maps based on the extracted features, $D(\vec{a_i}) \rightarrow \hat{\vec{y_i}} \in \Re^{n}$. In this section, we first present the CNN architectures used in our EML-NET. Secondly, we show the procedures of training an encoder and a decoder in a space efficient way. 

\subsection{Convolutional Neural Networks}\label{sec:cnn}

The VGGNet-16 and ResNet-50 models are applied in most of the state-of-the-art saliency models \cite{DeepGaze2,DeepFix,DSCLRCN,SAMres,Wang17a}. Specifically, ResNet-50 achieves a higher accuracy than VGGNet-16 \cite{ResNet} on ImageNet for object classification due to more model parameters. It has also been proved that ResNet-50 outperforms VGGNet-16 for saliency prediction \cite{DSCLRCN,SAMres}. In our experiment, we further explore the even deeper CNN models, DenseNet-161 \cite{DenseNet} and NasNet-Large \cite{NasNet}, for the study of saliency.

The DenseNet architecture contains several convolutional blocks, denoted as dense blocks. Instead of using skip connections (used in ResNet), a convolutional layer connects all the following layers within each dense block. The output of each dense block is concatenated with the input maps before passing to the next block. The model applied in this paper is DenseNet-161, which consists of four dense blocks, $160$ convolutional layers in total. Each of the blocks is followed by a transition layer for average pooling. The pooled features combined with the output of the last convolutional layer are extracted for multi-layer features.

More recently, Barret Zoph et al. proposed to use a recurrent network as a controller to automatically design CNN architectures. In their work, they discovered that basic units on the CIFAR dataset \cite{CIFAR}, defined as normal and reduction cells, can generalize well on other datasets. We apply the architecture of NasNet-Large (NasNet-A 6@4023) to learn saliency features, in which a reduction cell is added after every six normal cells. The output of the two reduction cells is extracted and combined with the multi-layer features from the DenseNet-161 model. 

In this work, not only are multi-layer features extracted from different CNN models for saliency prediction, but also we try to utilize prior knowledge learned from various vision tasks. The DenseNet-161 model is pre-trained on the PLACE365 dataset while the NasNet-Large model is pre-trained on ImageNet. 

\subsection{Encoding Stage}\label{sec:encoder}
The use of deep CNN models requires massive memory for training and this becomes even more problematic when combining multi-level features from different models. For the sake of scalability, we try to train our system in an almost end-to-end \emph{piecewise} fashion; the encoder and decoder are separately trained. Furthermore, at this stage, each CNN model (NasNet and DenseNet) is also separately optimized for saliency prediction.

The applied CNN models have been pre-trained for the task of image classification. In order to predict saliency maps, the Fully Connected (FC) layer is replaced with convolutional layers. This change is widely used in various dense labeling tasks, e.g. semantic segmentation and salient object detection. The number of parameters of FC layers grows linearly with the size of input images therefore space can be greatly saved by removing FC layers. But the parameters of the decoder still scale with the number of connected maps, which can be a bottleneck when utilizing multi-level features or multiple models. Let's take the NasNet model as an example, there are $568$ convolutional layers and one FC layer. The last convolutional layer contains $4,032$ filter maps, two times as many as the number used in ResNet-50, \cite{SAMres,DSCLRCN}. It demands even more space when utilizing features from multiple layers, e.g. $4,032+2,688+1,344=8,064$ maps for NasNet in our work.

To relax the requirement on space, the output prediction is compressed into one feature map by applying conv1@1 (kernel size @ number of filter maps). When computing losses for back-propagation, the output map is resized to the same size as the input by using bi-linear up-sampling. Multi-level features are not extracted at this stage in order to save space. The number of parameters of the conv1@1 operation are $4,032$ for NasNet and $2,208$ for DenseNet, the same number as the input filter maps. The ReLU function is applied after each convolution layer.

\subsection{Decoding Stage}\label{sec:decoder}
At this stage, we train a decoder to combine the learned features from the two CNN models that have been trained at the encoding stage. Moreover, the combined feature is from multiple layers as described in Section~\ref{sec:cnn}, four layers from the DenseNet model and three from NasNet, totalling seven layers, $13,536$ feature maps. Due to the size of the CNN models, it is costly to directly apply convolutional filters on the concatenated layer as done in \cite{MLNet}. It also requires much space to store the intermediate maps without extra information gain when aligning the size of multi-level features to the input \cite{DeepGaze1}. 

In our design, we first compress each of the selected layers into one feature map by applying conv1@1 and ReLU. When aligning and concatenating the seven maps, we only resize them to the size of the largest feature map instead of the input. In this work, given an input size $640 \times 480$, the largest size of the intermediate features is $160 \times 120$ from the DenseNet model. That is we only align seven feature maps to $160 \times 120$ by bilinear up-sampling. Our design can significantly save space for storage, three times less than aligning to the size of the input, about $19$ hundred times less than up-sampling the total $13,536$ feature maps (bilinear up-sampling can not deliver extra information but requires more space). When training a decoder to combine the multi-level features, the weights of the CNN models are frozen so that the size of the models can be halved due to no gradients being required, as shown in the blue box in Figure~\ref{fig:eml}.

The number of parameters of the decoder is $13,536+7=13,543$, $18$ hundred times less than the ResNet-50 model on ImageNet images ($224 \times 224$). Each CNN model is trained separately at the encoding stage therefore the total required computational space for EML-NET depends on the largest single CNN model, NasNet in this paper. Therefore, our EML-NET can be further expanded by combining CNN models with prior knowledge from other tasks.

\subsection{Saliency Metrics}
In this section, we show the saliency metrics used for our EML-NET and introduce the modified and combined loss function. Let's first denote a predicted saliency map as $P$ and the ground truth of density map as $Q$, these two can be considered as distributions. The ground truth fixation map only contains binary values, {0,1}, denoted as $F$, representing the location of fixations. 

\paragraph{Kullback-Leibler Divergence (KLD)}
KLD measures the similarity between two distributions, density maps in this case. We show the equation implemented in the works \cite{Zoya16,SAMres}:

\begin{equation}
KLD(P, Q) = \sum_{i}Q_i log(\epsilon+\frac{Q_i}{P_i+\epsilon})
\end{equation}
where $i$ denotes the location of pixels in a saliency map and $\epsilon$ is a regularization term.

\paragraph{Pearson's Correlation Coefficient (CC)}
CC is a statistical method that measures the linear correlation between two distributions.
\begin{equation}
CC(P, Q) = \frac{\sigma(P,Q)}{\sigma(P) \times \sigma(Q)}
\end{equation}
where $\sigma(P,Q)$ denotes the covariance of $P$ and $Q$.

\paragraph{Normalized Scanpath Saliency(NSS)}
NSS was used to measure the average normalized saliency between two fixation maps, $P$ and $F$ in \cite{Peters05}. 
\begin{equation}
\begin{array}{c}
NSS(P, F) = \frac{1}{N}\sum_{i}\bar{P_i} \times F_i \\
\text{where} \quad N = \sum_{i} F_i \quad \text{and} \quad \bar{P} = \frac{P - \mu(P)}{\sigma(P)} 
\end{array}
\end{equation}
$\mu(\cdot)$ and $\sigma(\cdot)$ respectively represent the mean and standard deviation of the input.

These three saliency metrics are combined to formulate our loss function, but we report results using more saliency metrics, see Section~\ref{sec:exp3}. Each saliency metric focuses on different attributes \cite{Zoya16}. For instance, AUC metrics place more attention on FPs with high values but low-valued FPs will be ignored, KL is more sensitive to FNs, CC and NSS equally penalize FPs and FNs \footnote{AUC: Area Under Curve, FP:False Positive, FN:False Negative}. 

\subsection{Combined Loss Function}\label{sec:loss}
Not all saliency metrics perform similarly to the cross-entropy loss, e.g. the range of CC is $[-1, 1]$ and $1$ represents the two distributions are identical, a higher NSS score indicates a better prediction is. To combine the three metrics, we first modify the metirc of CC by:

\begin{equation}
CC'(P, Q) = 1 - \frac{\sigma(P,Q)}{\sigma(P) \times \sigma(Q)}
\end{equation}
We use $CC'$ to represent our modified CC metric. This $CC'$ converts the similarity metric into dissimilarity in the range of $[0, 2]$. A value of zero means the two distributions are perfectly matched. 

For the NSS metric, the modified version is shown as:
\begin{equation}
\begin{array}{c}
NSS'(P, F) = \frac{1}{N}\sum_{i}(\bar{R_i} - \bar{P_i}) \times F_i \\
\text{where} \quad N = \sum_{i} F_i \quad \bar{P} = \frac{P - \mu(P)}{\sigma(P)} \\
\text{and} \quad \bar{R} = \frac{F - \mu(F)}{\sigma(F)}
\end{array}
\end{equation}
The $\bar{P_i}$ and $\bar{R_i}$ represent normalized predictions and ground truth. Imagine that the prediction map $P_i$ shrinks to discrete fixation points and its locations are perfectly matched with the ground truth, the loss would be zero. In practice, the prediction map is a gaussian-like distribution where low values can be considered as FNs when computing NSS. Which may result in a lower value of $\bar{P_i}$, a higher loss of $NSS'$. We simply take the summation of $KLD$, $CC'$ and $NSS'$ to formulate our loss function, $Loss = NSS' + CC' + KLD$. Note that the modified metrics are only used for training, the original metrics are used to measure and report results on the benchmarks.

\section{Experiments}\label{sec:results}

\begin{table*}[t]
\renewcommand{\arraystretch}{1.3}

\centering
\begin{tabular}{|c|c c c c c c |}
\hline
Model & NSS& CC&AUC&sAUC&KLD& SIM\\
\hline
DenseNet & 1.987&0.873&0.797&0.774&0.230&0.769\\
NasNet & 1.996&0.884&0.803&0.780&0.216&0.777\\
NasNet + DenseNet & 2.024&0.890&0.802&0.778&0.204&0.785\\
\hline
\end{tabular}
\caption{Performance of different models on the SALICON validation set.}
\label{tab:val}
\end{table*}

\begin{figure*}[!tph]
\centering
\includegraphics[width=.98\textwidth]{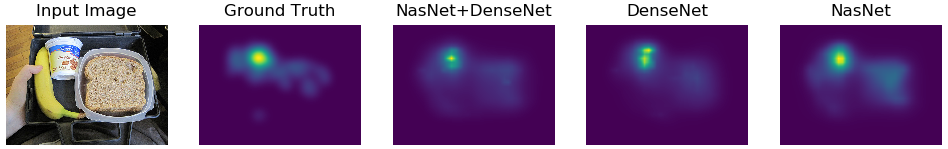}\\
\includegraphics[width=.98\textwidth]{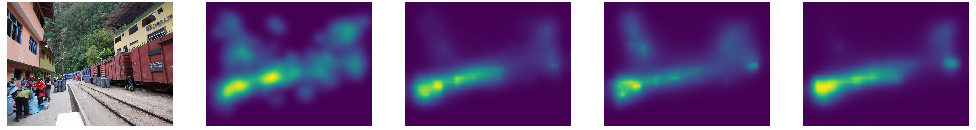}\\
\includegraphics[width=.98\textwidth]{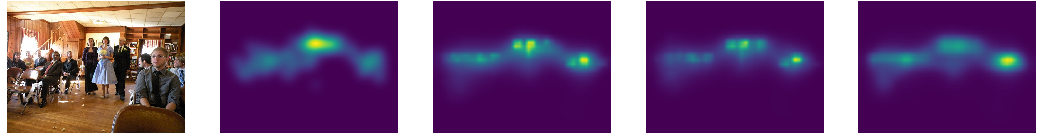}\\
\includegraphics[width=.98\textwidth]{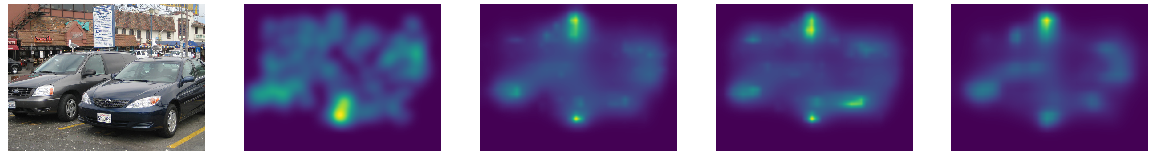}\\
\caption{Examples on the SALICON validation set.}
\label{fig:exa_results}
\end{figure*}

\subsection{Saliency Datasets}
\paragraph{SALICON:} 
The Saliency in Context dataset \cite{SALICONdb} contains $10,000$ images for training, $5,000$ images in the validation set and $5,000$ for test. Those images were collected from the MS-COCO  challenge \cite{COCO} and the ground truth was labelled based on mouse-tracking. They have shown correlation between mouse and eye tracking data for saliency annotation. The SALICON dataset is the largest for saliency prediction, which enables the exploration of deep CNN models. Our experiments were based on the newest release, SALICON 2017, from the Large-Scale Scene Understanding challenge.

\paragraph{MIT300:}
The MIT300 dataset was built similar to MIT1003, from the same sources and annotation procedures. But the ground truth was collected using eye-tracking from $39$ observers, and the labels are non-public. We evaluate our proposed EML-NET on this dataset to compare against other state-of-the-art saliency methods.

\paragraph{MIT1003:}
The MIT1003 dataset contains $1,003$ images collected from Flickr and LabelMe. The ground truth of MIT1003 was annotated by eye tracking devices based on $15$ observers. As suggested in MIT300, this dataset is used to fine-tune our EML-NET.

\paragraph{CAT2000:}
The CAT2000 dataset contains $4,000$ images for saliency prediction, $2,000$ for training and $2,000$ for test. Again, the training images were further split into $1,800$ for fine-tuning our EML-NET and $200$ for validation. This dataset is more challenging to our EML-NET because some of the image styles are not common in image classification, e.g. cartoon, art, line drawing  satellite or even jumbled, see Section~\ref{sec:exp3} for more discussion.

\begin{table*}[t]
\renewcommand{\arraystretch}{1.3}

\centering
\begin{tabular}{|c|c c c c c c c|}
\hline
Rank & NSS& CC&AUC&sAUC&KLD& IG & SIM\\
\hline
EML-NET & \textcolor{red}{2.050}&\textcolor{blue}{0.886}&\textcolor{red}{0.866}&\textcolor{red}{0.746}&\textcolor{blue}{0.520}&\textcolor{red}{0.736}&\textcolor{blue}{0.780}\\
Second place &1.928&0.874&0.862&\textcolor{blue}{0.745}&\textcolor{red}{0.489}&\textcolor{blue}{0.646}&0.771\\
Third place &\textcolor{blue}{2.045}&0.862&0.862&\textcolor{blue}{0.745}&1.026&0.357&0.753\\
SAM-Res\cite{SAMres}&1.990&\textcolor{red}{0.899}&\textcolor{blue}{0.865}&0.741&0.610&0.538  & \textcolor{red}{0.793}\\
\hline
\end{tabular}
\caption{Performances of EML-NET on the SALICON test set.}
\label{tab:salicon}
\end{table*}

\begin{table*}[!pth]
\renewcommand{\arraystretch}{1.3}

\centering
\begin{tabular}{|c|c c c c c c c c|}
\hline
Method & AUC-Judd& SIM&EMD&AUC-Borji &sAUC&CC& NSS & KLD\\
\hline
eDN\cite{EDN} & 0.82&0.41&4.56&0.81&0.62&0.45&1.14&1.14\\
DeepGaze1\cite{DeepGaze1} & 0.84&0.39&4.97&0.83&0.66&0.48&1.22&1.23\\
DeepGaze2\cite{DeepGaze2} & \textcolor{red}{0.88}&0.46&3.98&\textcolor{red}{0.86}&0.72&0.52&1.29&0.96\\
BMS\cite{BMS} &0.83&0.51&3.35&0.82&0.65&0.55&1.41&0.81\\
iSEEL\cite{iSEEL} &0.84&0.57&2.72&0.81&0.68&0.65&1.78&0.65\\
DVA\cite{Wang17a} &0.85&0.58&3.06&0.78&0.71&0.68&1.98&0.64\\
SalGAN\cite{SalGAN} &0.86&0.63&2.29&0.81&0.72&0.73&2.04&1.07\\
PDP\cite{PDP} & 0.85&0.60&2.58&0.80&0.73&0.70&2.05&0.92\\
ML-Net\cite{MLNet} & 0.85&0.59&2.63&0.75&0.70&0.67&2.05&1.10\\
Salicon\cite{Salicon} &0.87&0.60&2.62&\textcolor{blue}{0.85}&\textcolor{red}{0.74}&0.74&2.12&\textcolor{red}{0.54}\\
DeepFix\cite{DeepFix} &0.87&0.67&\textcolor{blue}{2.04}&0.80&0.71&0.78&2.26&\textcolor{blue}{0.63}\\
SAM-Res\cite{SAMres} &0.87&\textcolor{blue}{0.68}&2.15&0.78&0.70&0.78&2.34&1.27\\
DSCLRCN\cite{DSCLRCN} & 0.87&\textcolor{blue}{0.68}&2.17&0.79&0.72&\textcolor{blue}{0.80}&2.35&0.95\\
DPN\cite{DPN} & 0.87&\textcolor{red}{0.69}&2.05&0.80&\textcolor{red}{0.74}&\textcolor{red}{0.82}&\textcolor{blue}{2.41}&0.91\\
\hline
EML-NET & \textcolor{red}{0.88}&\textcolor{blue}{0.68}&\textcolor{red}{1.84}&0.77&0.70&0.79&\textcolor{red}{2.47}&0.84\\
\hline
\end{tabular}
\caption{Performances of EML-NET on the MIT300 dataset.}
\label{tab:mit300}
\end{table*}

\subsection{Ablation Study of Architectures}\label{sec:exp1}
In the work of \cite{Salicon}, VGGNet-16 outperforms the deeper model, GoogleNet, when training on $450$ images. However, the dataset set they used to train the huge model is arguably too small. Recently, state-of-the-art saliency models \cite{DSCLRCN,SAMres} were built on the large-scale SALICON dataset. ResNet-50 achieved higher performances than the VGGNet-16 model in these works. In our experiments, we first compared the two CNN models and report results on the SALICON validation set.

Each model was first trained on the SALICON training set using the encoding protocol, we only output saliency maps from the last layer. The input image size was $640 \times 480$ and the size of batch was eight. The initial learning rate was $0.1$ and we decayed the rate by multiplying $0.1$ after five epochs. A momentum of $0.9$ and a weight decay of $1e-4$ were also applied. Secondly, we trained a decoder for each model to combine multi-level features as described in Section~\ref{sec:decoder}. The decoder was trained using a bigger batch size of $32$ for five epochs and the learning rate was multiplied by $0.1$ at the third epoch.

As shown in Table~\ref{tab:val}, the NasNet model outperforms DenseNet on all the metrics. This experiment proves that deeper CNN models are able to better extract saliency features when the training set is large enough. An even higher performance can be achieved by combining the two models. However, the NasNet model obtained higher performances than the combined on the AUC metrics. We believe the reason behind this is due to the bias of AUC metrics. As shown in Figure~\ref{fig:exa_results} the first row, for instance, less FPs with low values (noodle area) can not increase the performance in considering AUCs. Meanwhile, less FPs with high values near the most salient region results in a lower AUC score. Similarly, the second row of Figure~\ref{fig:exa_results} shows TPs with low values (the area of the building on the left) are also ignored by AUC metrics. The rest of Figure~\ref{fig:exa_results} also shows the different areas of attention drawn by the two models. The differences in prediction and improvement is likely a result of the prior knowledge learned from different visual tasks and our EML-NET, which can be easily expanded in merging more CNN models.

\subsection{Comparison of State-of-the-art}\label{sec:exp2}
To further validate our EML-NET, we report results on the public saliency benchmarks where the ground truth is unknown. Different saliency metrics were used to evaluate our method. 

We applied our EML-NET on the SALICON-2017 test set\footnote{\url{https://competitions.codalab.org/competitions/17136}}, $5,000$ maps were submitted to the challenge system. as shown in Table~\ref{tab:salicon}, our method achieved the first place on the leader board, the highest score on the metric NSS.

\begin{table*}[!pth]
\renewcommand{\arraystretch}{1.3}

\centering
\begin{tabular}{|c|c c c c c c c c|}
\hline
Method & AUC-Judd& SIM&EMD&AUC-Borji &sAUC&CC& NSS & KLD\\
\hline
eDN\cite{EDN} & 0.85&0.52&2.64&\textcolor{red}{0.84}&0.55&0.54&1.30&0.97\\
BMS\cite{BMS} & 0.85&0.61&1.95&\textcolor{red}{0.84}&\textcolor{red}{0.59}&0.67&1.67&0.83\\
iSEEL\cite{iSEEL} & 0.84&0.62&1.78&0.81&\textcolor{red}{0.59}&0.66&1.67&0.92\\
DeepFix\cite{DeepFix} &\textcolor{blue}{0.87}&\textcolor{blue}{0.74}&1.15&0.81&0.58&\textcolor{blue}{0.87}&2.28&\textcolor{red}{0.37}\\
SAM-Res\cite{SAMres} & \textcolor{red}{0.88}&\textcolor{red}{0.77}&\textcolor{red}{1.04}&0.80&0.58&\textcolor{red}{0.89}&\textcolor{red}{2.38}&\textcolor{blue}{0.56}\\
\hline
EML-NET & \textcolor{blue}{0.87}&\textcolor{blue}{0.74}&\textcolor{blue}{1.05}&0.78&0.58&\textcolor{blue}{0.87}&\textcolor{red}{2.38}&0.95\\
\hline
\end{tabular}
\caption{Performances of EML-NET on the CAT2000 dataset.}
\label{tab:cat2000}
\end{table*}

\begin{figure}[tph]
\centering
\includegraphics[width=0.46\textwidth]{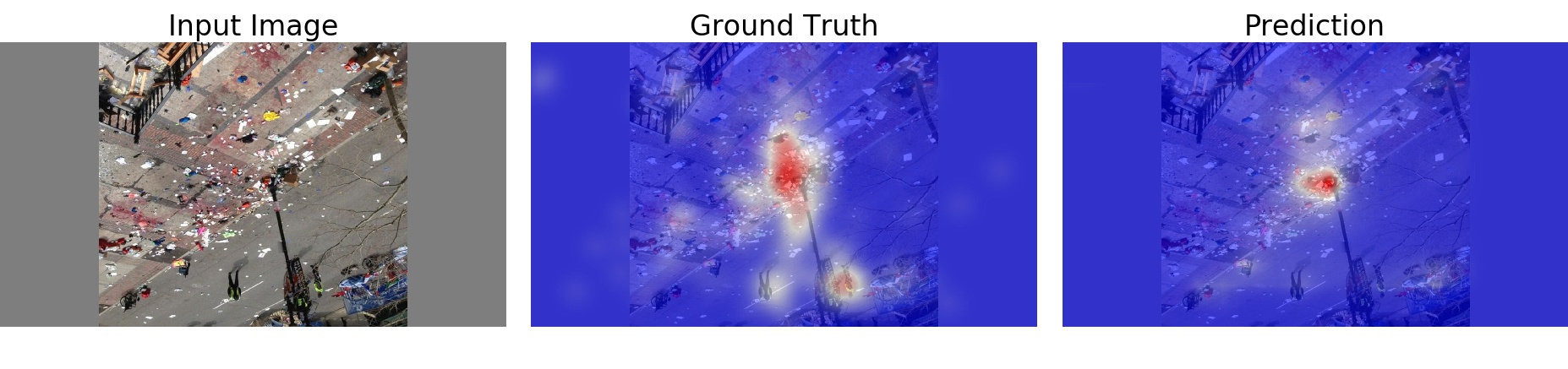}\\
\includegraphics[width=0.46\textwidth]{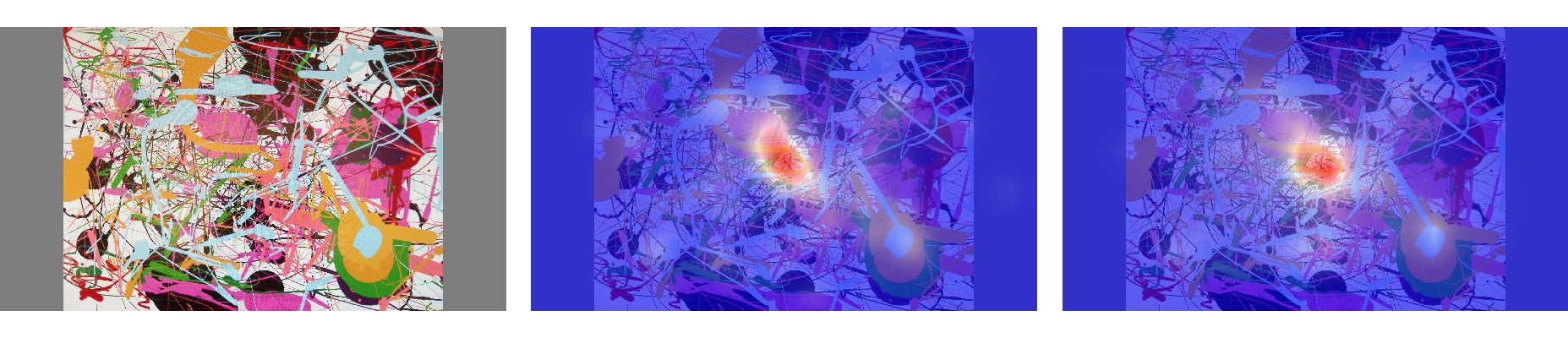}\\
\includegraphics[width=0.46\textwidth]{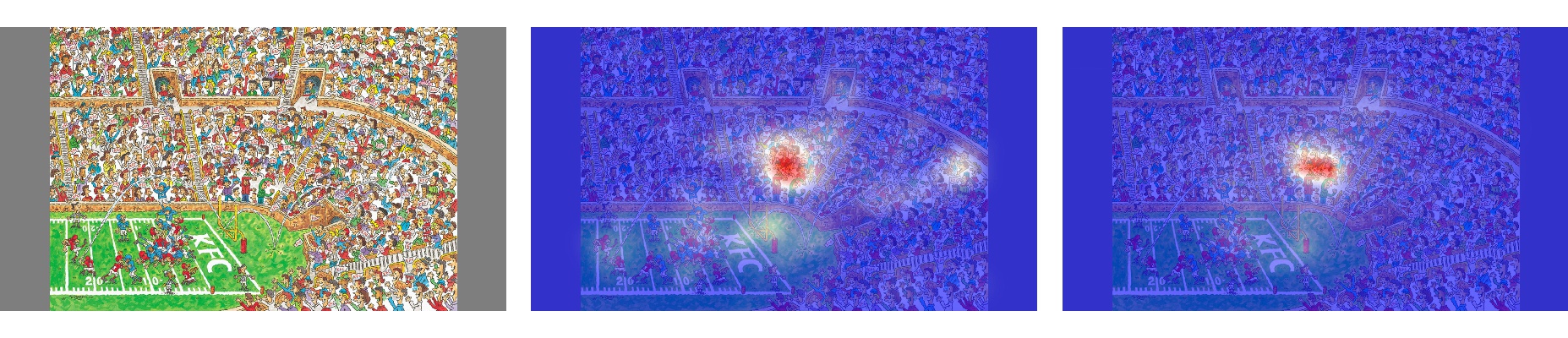}\\
\includegraphics[width=0.46\textwidth]{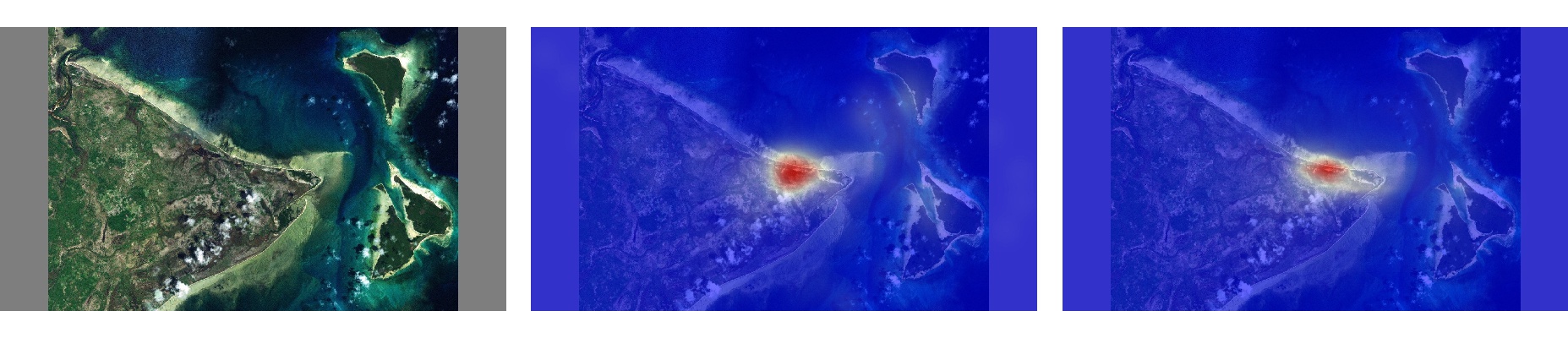}\\
\includegraphics[width=0.46\textwidth]{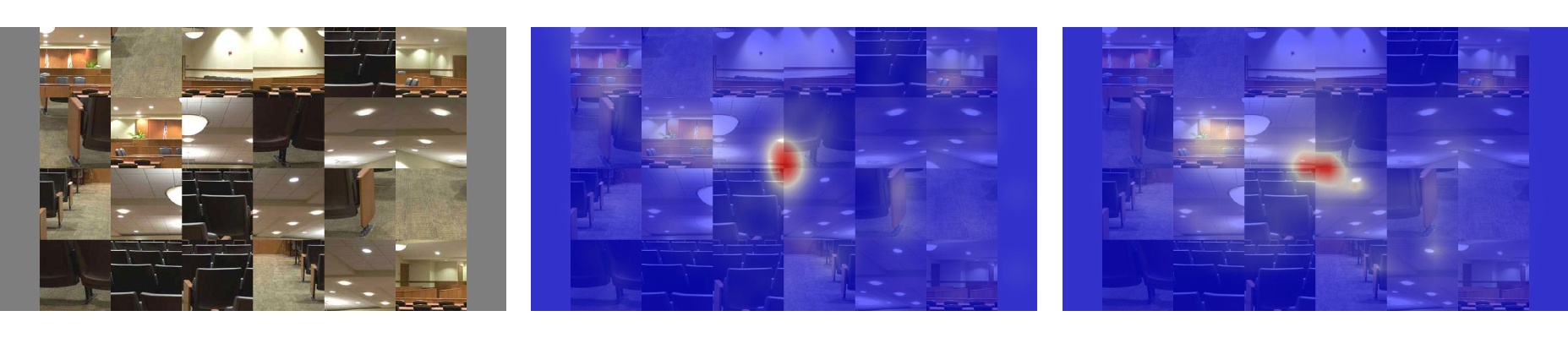}
\caption{Failure cases from the CAT2000 dataset. First row :`Inverted'; Second row : `Art'; Third row : `Cartoon'; Fourth row : `Satellite'; Fifth row : `Jumbled'.}
\label{fig:exa_cat}
\end{figure}

We further evaluated our EML-NET on the MIT300 benchmark. As suggested by the organizers \cite{MIT300a,MIT300b}, we fine-tuned our EML-NET on the MIT1003 dataset following the protocol used in \cite{SAMres,DeepFix}. The MIT1003 dataset was randomly split into training and test sets, containing $903$ and $100$ images respectively. We applied zero-padding on the images in order to align the ratio between the width and height to $4:3$. Then all the images were resized to the same size as SALICON, $640 \times 480$. We further fine-tuned the encoder from the last experiment (SALICON) on the MIT1003 training set applying all the settings in Section~\ref{sec:exp1}.

As suggested in \cite{Zoya16,MIT300a,MIT300b}, the leader board of MIT300 is ranked based on the metric of NSS and our EML-NET achieved the highest score, $2.47$. As shown in Table~\ref{tab:mit300}, we also achieved the lowest earth mover distance (lower is better) score, $1.84$, $0.2$ lower than the second place \cite{DeepFix}. Similarly, we also report results on the CAT2000 benchmark. $2,000$ images with annotations were split to $1,800$ for training and $200$ for validation. The input image was resized to $640 \times 480$ to output saliency maps and then the resulting map was resized to its original size, e.g. $1920 \times 1080$ for CAT2000. However, our EML-NET did not obtain higher performances on this dataset. As shown in Table~\ref{tab:cat2000}, the NSS score achieved is the same as \cite{SAMres}. The CAT2000 dataset is split into sub-categories and the metrics are separately computed on each category. We did a failure case study by checking those categories on which our model achieved low performance. The main reason appears to be that images in CAT2000 are very different from the natural images in ImageNet or PLACE365. As shown in Figure~\ref{fig:exa_cat}, some categories are really rare in common vision datasets, such as `Satellite', `Art', `Cartoon', `Inverted' and `Jumbled'. The ground truth suggests humans tend to gaze at the center of the image when there are no obvious objects. Therefore our EML-NET can not utilize prior knowledge for saliency prediction and could be over-fitting in this scenario due to the small training set for those categories. A potential solution to this problem could be combining a new encoder model which contains prior knowledge regarding those rare categories.

\section{Conclusion}
In this paper, we have shown deeper CNN models can deliver better saliency features when the training set is large enough. Moreover, we have presented a scalable method for combining many networks of arbitrary depth and complexity that may each encode non-overlapping information relevant to determining visual saliency. The proposed method has achieved the state-of-the-art result on the public saliency benchmarks. The main advantage of our EML-NET is scalability; computation can be more efficient in applying deeper CNN models or combining prior knowledge from other vision tasks, e.g. face detection or satellite image analysis. With the increasing availability of visual datasets, off-the-shelf models could be combined to expand the system on new categories, e.g. `Jumbled' in CAT2000. Furthermore, our model can be also combined with a recurrent network for potential further improvements.


{\small
\bibliographystyle{ieee}
\bibliography{egbib}
}

\end{document}